\documentclass[11pt]{article}

\usepackage[final]{acl}

\usepackage{times}
\usepackage{latexsym}

\usepackage[T1]{fontenc}
\usepackage[utf8]{inputenc}

\usepackage{microtype}

\usepackage{inconsolata}

\usepackage{hyperref}
\usepackage{url}
\usepackage{enumitem}
\usepackage{booktabs} 
\usepackage{multirow}
\usepackage{graphicx}
\usepackage{enumitem} % 在导言区添加
\usepackage[table]{xcolor}
\definecolor{mygray}{gray}{0.93} % 浅灰底色
\usepackage{graphicx}    % 插图必需
\usepackage{subcaption}  % 子图支持
\usepackage{wrapfig}
\usepackage{algorithm}
\usepackage{algorithmic}
\usepackage{amsmath}
\usepackage{amssymb}
\usepackage{colortbl}
\usepackage{xcolor}

\usepackage{enumitem}  % 在导言区添加
\newcommand{\ie}{\emph{i.e., }}
\newcommand{\eg}{\emph{e.g., }}

\newcommand{\tabref}[1]{Table~\ref{#1}}
\newcommand{\figuref}[1]{Figure~\ref{#1}}
\newcommand{\equref}[1]{Eq.~\ref{#1}}

\title{Controllable LLM Reasoning via Sparse Autoencoder-Based Steering}

\author{Yi Fang\textsuperscript{1,2},
	Wenjie Wang\textsuperscript{1}\thanks{Corresponding author.}, 
	Mingfeng Xue,  
	Boyi Deng\textsuperscript{1}, 
    Fengli Xu\textsuperscript{2,3},
    Dayiheng Liu,
	Fuli Feng\textsuperscript{1}\footnotemark[1]
 \\
	\textsuperscript{1}University of Science and Technology of China, \\
	\textsuperscript{2}Zhongguancun Academy, 
    \textsuperscript{3}Tsinghua University \\
}

\begin{document}
\maketitle

\begin{abstract}
Large Reasoning Models (LRMs) exhibit human-like cognitive reasoning strategies (\eg backtracking, cross-verification) during the reasoning process, which improves their performance on complex tasks. Currently, reasoning strategies are autonomously selected by LRMs themselves. However, such autonomous selection often produces inefficient or even erroneous reasoning paths. To make reasoning more reliable and flexible, it is important to develop methods for controlling reasoning strategies. Existing methods struggle to control fine-grained reasoning strategies due to conceptual entanglement in LRMs' hidden states. To address this, we leverage Sparse Autoencoders (SAEs) to decompose strategy-entangled hidden states into a disentangled feature space. To identify the few strategy-specific features from the vast pool of SAE features, we propose SAE-Steering, an efficient two-stage feature identification pipeline. SAE-Steering first recalls features that amplify the logits of strategy-specific keywords, filtering out over 99\% of features, and then ranks the remaining features by their control effectiveness. Using the identified strategy-specific features as control vectors, SAE-Steering outperforms existing methods by over 15\% in control effectiveness. Furthermore, controlling reasoning strategies can redirect LRMs from erroneous paths to correct ones, achieving a 7\% absolute accuracy improvement. Our code and data are available at \url{https://github.com/Peter-Fy/SAE-Steering}.
\end{abstract}

\section{Introduction}

\begin{figure}[t]
    \centering
    \includegraphics[width=\linewidth]{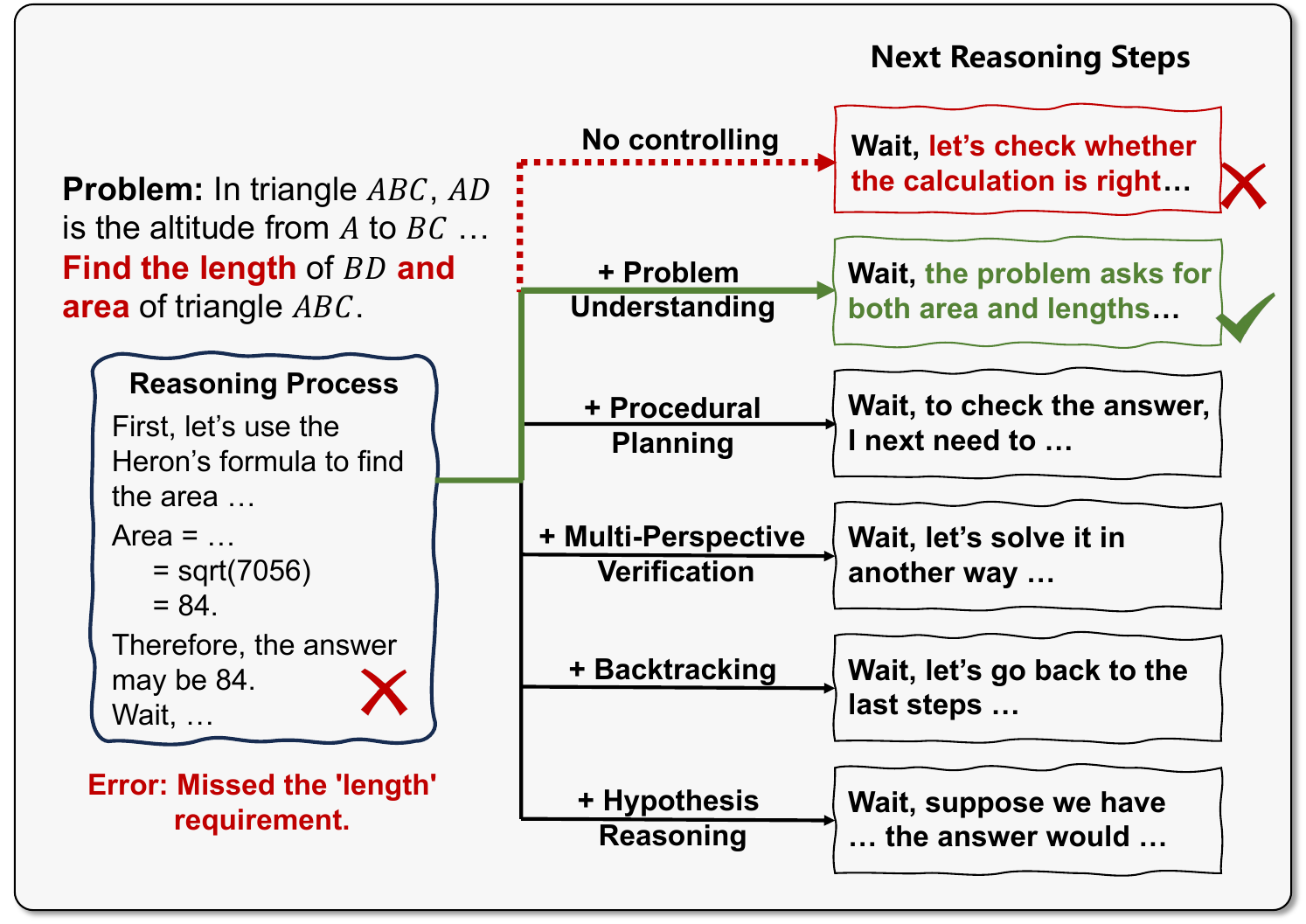}
    \caption{An illustration of reasoning strategy control. By deliberately controlling the LRM's strategy selection, we can flexibly intervene and correct its reasoning path when a flaw emerges.}
    \label{fig:overview}
\end{figure}

Large Reasoning Models (LRMs), such as GPT-o1~\citep{openai2025o3mini} and DeepSeek-R1~\citep{guo2025deepseek}, employ a ``think-then-answer'' paradigm, explicitly generating intermediate reasoning processes before deriving final answers.
Within these reasoning processes, LRMs exhibit human-like cognitive reasoning strategies such as self-correction and cross-verification~\citep{reasong_strategy,marjanović2025deepseekr1thoughtologyletsthink,pan2025surveyslowthinkingbasedreasoning}. Such reasoning strategies improve the accuracy and robustness of LRMs on challenging tasks~\citep{Snell0XK2025scaling,zaremba2025trading}.
These LRMs autonomously select reasoning strategies during reasoning. However, such autonomous reasoning often produces inefficient or even erroneous reasoning paths~\citep{overthinking,underthinking1}. To improve the reliability and flexibility of reasoning, external guidance is promising.
For example, as illustrated in Figure~\ref{fig:overview}, if an LRM misinterprets the problem but pursues a flawed verification path, external guidance can redirect it to re-examine the problem statement, correcting the error. Therefore, developing methods for deliberate control over reasoning strategies is crucial.

Existing control methods fall into two categories: prompt-based and activation-based. Prompt-based methods control the LRM's reasoning by incorporating instructions either in the initial user prompt~\citep{self_discover} or during intermediate reasoning stages~\citep{ThinkingIntervention,zhang-etal-2025-steer}.
However, these methods lack direct control over the LRM's internal generative process, which results in frequent instruction-following failures, especially when reasoning context is long or instructions conflict with pre-trained behaviors~\citep{instruction_follow}.
Activation-based methods offer more direct control by deriving a control vector to modify the LRM's hidden states during generation~\citep{SteeringVector_ICLR}. This control vector is typically computed as activation differences between contrastive pairs exhibiting or lacking a target behavior~\citep{resong_strength2}. 
However, curating contrastive pairs that cleanly isolate a single strategy is difficult. As a result, the derived control vectors are prone to concept entanglement~\citep{entangle1,entangle2}, inadvertently capturing features of multiple strategies and hindering precise control.

To overcome this limitation, we propose leveraging Sparse Autoencoders (SAEs)~\citep{SAE_sparsity} to decompose the LRM's hidden states into a sparse set of interpretable and monosemantic features~\citep{claudeTowards}. Specifically, a well-trained SAE projects the low-dimensional, strategy-entangled hidden states of an LRM into a high-dimensional, disentangled feature space. This projection aims to isolate strategy-specific features in the high-dimensional space, thereby providing disentangled control vectors for reasoning strategy control. However, the high-dimensional feature space introduces a new challenge: identifying the few strategy-specific features from tens of thousands of learned SAE features. Existing selection methods~\cite{SAEReasoning}, which rely on differential activation strength across contrastive pairs, face the same difficulty in constructing clean contrastive pairs. Furthermore, high activation does not guarantee effective control, leading to the selection of many spurious or ineffective features.

\begin{figure*}[tbh]
    \centering
    \includegraphics[width=0.95\linewidth]{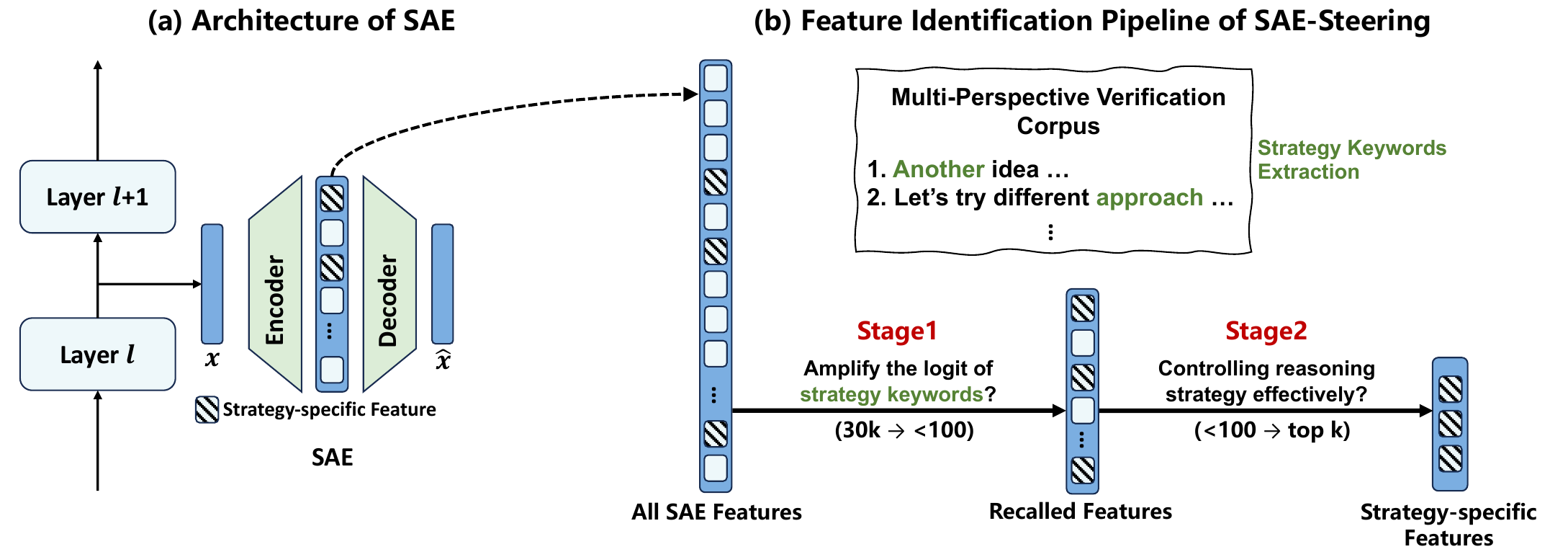}
    \caption{(a) Overview of the SAE architecture. (b) Feature identification pipeline of SAE-Steering. Numbers below the arrows indicate the approximate count of features retained. }
    \label{fig:pipeline}
\end{figure*}

To address this, we propose identifying effective features by directly assessing their capacity to steer target strategy generation. Considering exhaustively evaluating all features is computationally infeasible, we introduce \textbf{SAE-Steering}, a two-stage pipeline for efficiently identifying and selecting effective strategy control features, balancing cost and precision. 
As shown in Figure~\ref{fig:pipeline}, SAE-Steering first employs a low-cost, high-recall criterion to rapidly filter out over 99\% of irrelevant features by identifying those that amplify the logits of strategy-specific keywords---a strong indicator of control potential. It then applies a more computationally intensive evaluation to quantitatively assess and rank the control effectiveness of remaining candidates on a small validation set, selecting the most effective features for final application.
Extensive evaluations demonstrate that SAE-Steering consistently outperforms baselines by over 15\% in control effectiveness across various reasoning tasks and LRM architectures.
Moreover, SAE-Steering can correct erroneous reasoning paths in LRMs, improving absolute accuracy by 7\%, highlighting the potential of strategic control.
In summary, the contributions of this work are threefold:
\begin{itemize}[itemsep=1pt, topsep=2pt, parsep=0pt, leftmargin=*]
    \item We leverage SAEs to disentangle and identify strategy-specific features, overcoming the concept entanglement problem inherent in controlling reasoning strategies.
    \item We propose SAE-Steering to identify strategy‑specific features, addressing the challenge of efficient and effective feature selection from the massive set of SAE features. 
    \item Extensive experiments validate SAE-Steering's effectiveness and robustness in controlling reasoning strategies and demonstrate its potential use in correcting erroneous reasoning paths.
\end{itemize}

\section{Preliminary}
\label{sec:preliminary}

\paragraph{Strategy Selection.}
LRMs employ a diverse range of cognitive reasoning strategies during their reasoning processes, making a comprehensive evaluation of control over each one impractical. Therefore, we focus on five representative reasoning strategies that are frequent, effective, and widely studied in prior work~\citep{reasong_strategy,zhong2024achieving}. 
As illustrated in Figure~\ref{fig:overview}, the five strategies we selected are:
% \begin{itemize}[itemsep=1pt, topsep=2pt, parsep=0pt, leftmargin=*]
\begin{itemize}[itemsep=1pt, leftmargin=*]
    \item \textbf{Problem Understanding}: rephrasing the problem statement, clarifying its constraints and interpreting the given information.
    \item \textbf{Procedural Planning}: defining a sub-task or outlining a plan for the subsequent reasoning.
    \item \textbf{Backtracking}: identifying a mistake in previous reasoning and attempting to correct it or revert to a prior step. 
    \item \textbf{Multi-Perspective Verification}: verifying a conclusion by applying a different method or examining specific cases.
    \item \textbf{Hypothesis Reasoning}: making an assumption or posing a "what if" scenario to explore possibilities or test certain conditions. 
\end{itemize}

Importantly, this selection is purely for evaluation convenience; our method is general and applicable to control other reasoning strategies as well. 

\paragraph{Task Formulation.}
We next formalize the task of controlling reasoning strategies. 
In a standard autoregressive setting, an LRM generates the next token $y_t$ based on the prefix $Y_{<t} = \{y_1, \dots, y_{t-1}\}$. The LRM processes $Y_{<t}$ through its $L$ transformer layers, producing a sequence of residual stream activations $\{ \mathbf{x}_t^1, \mathbf{x}_t^2, \dots, \mathbf{x}_t^L \}$.
In vanilla decoding, these activations remain unmodified. 
Strategy control departs from this by injecting a control vector $\Delta\mathbf{x}^\ell$ at a specific layer $\ell$:
\begin{equation}
    \mathbf{x}'^{\,\ell}_t = \mathbf{x}^\ell_t + \alpha \cdot \Delta\mathbf{x}^\ell,
    \label{eq:steering}
\end{equation}
where $\alpha \in \mathbb{R}$ is a coefficient controlling the steering strength. The activation $\mathbf{x}'^{\,\ell}_t$ then replaces $\mathbf{x}^{\,\ell}_t$ and is propagated through the remaining layers, influencing the final generation. 
Importantly, the intervention is not applied only once at the final token of the prefix. Instead, during controlled generation, we apply the intervention at every subsequent decoding step for $T$ consecutive tokens. This yields a steered trajectory $Y'=\{y'_t, y'_{t+1}, \ldots, y'_{t+T-1}\}$.
% By repeating this intervention for $T$ consecutive tokens, the LRM produces a steered trajectory $Y' = \{y'_t, y'_{t+1}, \dots, y'_{t+T-1}\}$.
Given a pre-specified reasoning strategy $s$, the goal of reasoning strategy control is to construct $\Delta\mathbf{x}^\ell$ such that the steered trajectory $Y'$ exhibits the desired strategy $s$.

\section{Method}

This section details our method in two parts. First, we describe how we control reasoning strategies by manipulating strategy-specific features identified in the SAE (Section~\ref{sec:SAE}). Second, we introduce SAE-Steering, a two-stage pipeline developed to effectively identify these features from the vast SAE feature pool (Section~\ref{sec:steering}).

\subsection{Strategy Control with SAE Features}
\label{sec:SAE}
We train SAEs to disentangle and identify strategy‑specific features, which then serve as the control vectors for strategy control. As illustrated in \figuref{fig:pipeline}a, an SAE is an encoder--decoder architecture trained to represent an input activation as a sparse linear combination of learned feature directions. 
Given a residual stream activation $\mathbf{x} \in \mathbb{R}^N$, it encodes $\mathbf{x}$ into a sparse feature activation vector $\mathbf{z} \in \mathbb{R}^M$ ($M \gg N$) and reconstructs it as $\hat{\mathbf{x}}$:
\begin{align}
    \mathbf{z} &= \sigma\left(\mathbf{W}_{\mathrm{enc}} (\mathbf{x} - \mathbf{b}_{\mathrm{dec}}) + \mathbf{b}_{\mathrm{enc}}\right), \\
    \hat{\mathbf{x}} &= \mathbf{W}_{\mathrm{dec}} \mathbf{z} + \mathbf{b}_{\mathrm{dec}},
\end{align}
where $\mathbf{W}_{\mathrm{enc}} \in \mathbb{R}^{M \times N}$,
$\mathbf{b}_{\mathrm{enc}} \in \mathbb{R}^{M}$, 
$\mathbf{W}_{\mathrm{dec}} \in \mathbb{R}^{N \times M}$,
$\mathbf{b}_{\mathrm{dec}} \in \mathbb{R}^{N}$, 
and $\sigma$ is an activation function. 

The SAE is trained to satisfy a dual objective: (1) minimizing the reconstruction error $\|\mathbf{x} - \hat{\mathbf{x}}\|_2^2$ and (2) enforcing a sparsity restriction, which dictates that the reconstruction must be constructed from only a few active latent directions\footnote{We enforce sparsity via a Top-$K$ activation function, which only retains the $K$ largest activation values and sets the rest to zero, following~\cite{TopK}.}.
This training process enables the SAE to approximate $\mathbf{x}$ as a sparse linear combination of the decoder columns:
\begin{align}
    \mathbf{x} \approx \mathbf{b}_{\mathrm{dec}} + \sum_{i=1}^{M} z_i(\mathbf{x})\mathbf{f}_i 
\end{align}
where each column $\mathbf{f}_i$ of $\mathbf{W}_{\mathrm{dec}}$ corresponds to a disentangled and interpretable latent direction, which we refer to as a \emph{feature} throughout the paper. 
The scalar $z_i(\mathbf{x})$ is the $i$-th component of the activation vector $\mathbf{z}$, indicating the activation strength of each feature for the input $\mathbf{x}$.

A key benefit of this decomposition is that the sparsity objective encourages \emph{monosemanticity}~\citep{claudeTowards}: each learned feature tends to capture a single concept, significantly mitigating the concept entanglement~\citep{SAE_sparsity}.

We then identify the strategy-specific feature $\mathbf{f}_s$ (one of the learned $\mathbf{f}_i$ directions) that is associated with the target reasoning strategy $s$ (see identification methods in Section~\ref{sec:steering}). By using $\mathbf{f}_s$ as the control vector $\Delta\mathbf{x}$ in \equref{eq:steering}, we steer the LRM's reasoning strategy by repeatedly injecting $\mathbf{f}_s$ into the residual stream activations at the SAE-trained layer $\ell$ for the next $T$ tokens generation:
\begin{align}
    \mathbf{x}'^{\,\ell}_{t+k} = \mathbf{x}^\ell_{t+k} + \alpha \cdot \mathbf{f}_s, \quad k = 0, 1, \dots, T-1
    \label{eq:steering_fs}
\end{align}
where $\alpha$ is the steering strength. The selection of $\alpha$ is a trade-off: excessively large values cause repetitive outputs~\citep{fu2021theoretical}, while excessively small values fail to control effectively. 
For each feature, we determine $\alpha$ by searching downwards from an empirically chosen high value, iteratively decreasing it until repetitive generation is eliminated (see Appendix~\ref{app:steering_strength} for details).

\subsection{Identification of Strategy-specific Features} 
\label{sec:steering}
To efficiently identify the few critical, strategy-specific features from tens of thousands of learned SAE features, we introduce SAE-Steering, a two-stage pipeline designed for both efficiency and precision. The first stage employs a low-cost, high-recall criterion to rapidly construct a compact candidate set, while the second stage applies a more computationally intensive, high-fidelity evaluation to select the most effective features.
As shown in Figure~\ref{fig:pipeline}b, SAE-Steering first recalls features that amplify the logits of strategy-specific keywords. This stage is low-cost and highly-efficient, filtering out 99\% of irrelevant features. Subsequently, SAE-Steering evaluates and ranks the control effectiveness of remaining candidates on a small validation set, selecting the top-ranked feature for application.

\paragraph{Stage 1: Recall based on logit estimation.} In the first stage, we efficiently distill a small set of promising candidates from tens of thousands of SAE features by selecting those that positively influence the logits of strategy keywords. The guiding hypothesis is that features which substantially increase these keyword logits are more likely to steer the LRM toward the corresponding reasoning strategy. 

Specifically, we first extract strategy keywords following the approach of~\citet{SAEReasoning}. These keywords serve as a computationally efficient proxy to identify features potentially correlated with the target strategy. Briefly, we first create a strategy-specific corpus by manually identifying reasoning segments in the LRM's responses. We then extract the most frequent words from each corpus to serve as strategy keywords (see Appendix~\ref{app:keywords} for the keywords list and identification details).

Next, we estimate all SAE features' potential logit contribution to strategy keywords using logit lens~\citep{logitlens}. Logit lens is a method commonly used to estimate the logit contribution of hidden state activations to each token in the vocabulary. We adapt it to SAE features as follows:

Formally, let $\mathbf{U} \in \mathbb{R}^{N \times V}$ be the LRM's unembedding matrix (\ie the weight matrix of the LM head), mapping hidden activations to logits over a vocabulary of size $V$.
Let $\mathbf{W}_{\mathrm{dec}} \in \mathbb{R}^{N \times M}$ be the SAE decoder matrix. As described in Section~\ref{sec:SAE}, each column of $\mathbf{W}_{\mathrm{dec}}$ corresponds to a disentangled feature direction $\mathbf{f}_i \in \mathbb{R}^{N}$.
We compute the logit contribution matrix $\mathbf{L} \in \mathbb{R}^{M \times V}$ for all features via:
\begin{align}
    \mathbf{L} = \mathbf{W}_{\mathrm{dec}}^{\top}\mathbf{U},
\end{align}
where the $i$-th row $\mathbf{L}_{i,:}$ gives the logit contributions of feature $\mathbf{f}_i$ across the vocabulary.
This computation requires only a single matrix multiplication, making it low-cost and efficient.

We aim to recall features that specifically and significantly amplify strategy keywords, while avoiding those that amplify irrelevant tokens more strongly than the keywords. To achieve this, we extract the top-10 tokens with the highest logit contribution for each feature and recall features satisfying: (i) at least $n$ of these tokens are strategy keywords, and (ii) each such keyword’s logit contribution exceeds a threshold~$\tau$. This recall step is highly selective, narrowing the candidate pool from tens of thousands of features to several tens.

\paragraph{Stage 2: Rank based on Control Effectiveness.} In the second stage, we evaluate and rank the candidate features from Stage 1 to identify those with the highest control effectiveness. This ranking is based on their empirical performance on a small validation set $\mathcal{P}$.

Formally, for each problem $p \in \mathcal{P}$ with a given response prefix $Y_{<t}$, we generate two distinct $T$-token continuations\footnote{We set the sampling temperature to 0 to eliminate randomness as a confounding factor in our evaluation.}: (i) a baseline trajectory $Y_0$, generated via standard decoding, and (ii) a steered trajectory $Y^{(j)}$, generated using the candidate feature $\mathbf{f}_j$ as the control vector. An LLM judge then assesses whether $Y^{(j)}$ more explicitly demonstrates the target strategy $s$ than $Y_0$~\footnote{We provide the prompt and validate the reliability of LLM Judges in Appendix~\ref{app:llm_judge_reliable}.}, yielding binary judgment $J_{p,j} \in \{0, 1\}$. The control effectiveness of a feature $\mathbf{f}_j$ is then calculated as the control success rate over the validation set:
\begin{align}
    \text{Effectiveness}(\mathbf{f}_j) = \frac{1}{|\mathcal{P}|}\sum_{p \in \mathcal{P}} J_{p,j}.
\end{align}
This empirical ranking allows us to select the top-ranked feature as $\mathbf{f}_s$ for the target strategy $s$.

\section{Experiments}

In this section, we conduct experiments to address the following research question:
\begin{itemize}[itemsep=1pt, topsep=2pt, parsep=0pt, leftmargin=*]
    \item \textbf{RQ1:} Can our SAE-based steering method, leveraging the identified features, reliably control LRMs' reasoning strategies?
    \item \textbf{RQ2:} How effective is SAE-Steering for strategy-specific feature identification?
    \item \textbf{RQ3:} Can we correct an LRM's erroneous reasoning path by deliberately controlling its reasoning strategies?
\end{itemize}

\subsection{Experiment Setup}
\label{sec:experiment_setup}

\paragraph{Datasets.} We train our SAEs on activations from a mixed corpus combining \textsc{LMSYS-Chat-1M}~\citep{ZhengLmsys} and \textsc{OpenThoughts-114K}~\citep{guha2025openthoughtsdatarecipesreasoning}, following prior work~\citep{SAEReasoning}. For the evaluation of reasoning strategy control, we first randomly sample 50 responses from past AIME competitions (1983–2023)~\citep{AIME} as the validation set. We then evaluate control effectiveness on 200 randomly sampled responses from AIME'24 and 25~\citep{AIME} and 200 responses from GPQA~\citep{gpqa}. GPQA is a science reasoning dataset spanning biology, physics, and chemistry, which we use to assess the out-of-domain generalization capability of our strategy-specific features.

\paragraph{Baselines.} We compare SAE-Steering with three representative control methods: 
\begin{itemize}[itemsep=1pt, topsep=2pt, parsep=0pt, leftmargin=*, label=$\circ$]
    \item \textbf{Logit Boosting}, which directly boosts the logits of strategy-specific keywords;
    \item \textbf{Think Intervention}~\citep{ThinkingIntervention}, which inserts human-crafted instructions into the middle of the reasoning process;
    \item \textbf{Vector Steering}~\citep{SteeringVector_ICLR}, which uses an LLM to annotate reasoning strategies for constructing contrastive datasets, then extracts control vectors via contrast pairs.
\end{itemize}

\paragraph{Evaluation Protocol.} We evaluate control effectiveness following the procedure described in Stage 2 of Section~\ref{sec:steering}. 
% Briefly, we generate the next 512 tokens of intermediate reasoning steps with and without feature steering. We then employ LLM judges to assess whether the target strategy is more explicitly and significantly demonstrated after steering. 
Importantly, for feature selection in Stage 2 of SAE-Steering, we use only GPT-4o~\citep{gpt4o} as the judge. For test evaluation, we employ three LLM judges---GPT-4o~\citep{gpt4o}, Gemini-2.5-flash~\citep{gemini}, and Deepseek-V3.2~\citep{deepseek}---to vote as judges.
This majority voting mitigates individual judge biases and ensures more reliable evaluation. We also test the agreement between LLM judges and human annotators, which achieves a high agreement rate of 0.82 (see Appendix~\ref{app:llm_judge_reliable} for details), confirming the reliability of LLM judges. The standardized prompt used for all LLM-as-a-Judge evaluations is shown in Appendix Figure~\ref{fig:llm_as_judges}.

\begin{table*}[tbh]
\centering
\resizebox{\textwidth}{!}{%
\begin{tabular}{ll|ccccc|ccccc|c}
\toprule
\multirow{2}{*}{Dataset} & \multirow{2}{*}{Method} & \multicolumn{5}{c|}{R1-Llama-8B} & \multicolumn{5}{c|}{Qwen3-8B} & \multirow{2}{*}{Average} \\
\cmidrule(lr){3-7} \cmidrule(lr){8-12}
& & PU & PP & BK & MV & HR & PU & PP & BK & MV & HR & \\
\midrule
\multirow{4}{*}{AIME} 
& Logit Boosting & 0.21 & 0.49 & 0.30 & 0.27 & 0.32 & 0.44 & 0.61 & 0.39 & 0.49 & 0.56 & 0.41 \\
& Think Intervention & 0.56 & 0.49 & 0.21 & 0.21 & 0.39 & 0.62 & 0.81 & 0.12 & 0.23 & 0.61 & 0.43 \\
& Vector Steering & 0.69 & 0.82 & 0.67 & 0.48 & 0.34 & 0.74 & 0.85 & 0.55 & 0.51 & 0.52 & 0.62 \\
\rowcolor{gray!20}
& SAE-Steering & \textbf{0.88} & \textbf{0.86} & \textbf{0.69} & \textbf{0.76} & \textbf{0.41} & \textbf{0.92} & \textbf{0.92} & \textbf{0.78} & \textbf{0.70} & \textbf{0.65} & \textbf{0.76} \\
\midrule
\multirow{4}{*}{GPQA} 
& Logit Boosting & 0.28 & 0.68 & 0.29 & 0.39 & 0.56 & 0.43 & 0.79 & 0.40 & 0.47 & 0.63 & 0.49 \\
& Think Intervention & 0.66 & 0.69 & 0.35 & 0.23 & 0.57 & 0.68 & 0.83 & 0.17 & 0.16 & 0.77 & 0.51 \\
& Vector Steering & 0.77 &\textbf{0.90} & 0.61 & 0.52 & 0.51 & 0.89 & 0.89 & 0.80 & 0.55 & 0.72 & 0.72 \\
\rowcolor{gray!20}
& SAE-Steering & \textbf{0.94} & \textbf{0.90} & \textbf{0.78} & \textbf{0.93} & \textbf{0.70} & \textbf{0.94} & \textbf{0.95} & \textbf{0.81} & \textbf{0.82} & \textbf{0.89} & \textbf{0.87} \\
\bottomrule
\end{tabular}%
}
\caption{Control effectiveness evaluation across five reasoning strategies: Problem Understanding (PU), Procedural Planning (PP), Backtracking (BK), Multi-Perspective Verification (MV), and Hypothesis Reasoning (HR). All results are based on majority voting among three LLM judges. We additionally report a robustness check excluding GPT-4o from the judge pool with 95\% confidence intervals in Appendix Table~\ref{tab:judge_robustness_results}.}
\label{tab:combined_results}
\end{table*}

\paragraph{Implementation Details.} 
We train TopK-SAEs~\citep{TopK} with latent dimension $M=65{,}536$ (\ie an expansion factor of 16 over
the LRM activation size $N=4{,}096$) and $K=50$ on the last layer of DeepSeek-R1-Distill-Llama-8B~\citep{guo2025deepseek} (hereafter referred to as R1-Llama-8B) and Qwen3-8B~\citep{yang2025qwen3technicalreport}.
For SAE-Steering hyperparameters, we set $n=2$ and $\tau=0.1$ in Stage 1, and continuation length $T=512$ in Stage 2. In all experiments, we apply only one strategy-specific
feature per test case; we do not simultaneously activate
multiple strategy features. For sampling, we set the temperature to 0 during control effectiveness evaluations to eliminate confounding effects from sampling stochasticity. For error correction experiments, we adopt the officially recommended temperature of 0.6 and set the maximum token length to 32,768. We additionally report inference latency on GPQA in Appendix Table~\ref{tab:latency} to provide transparency on the computational overhead of steering.

\subsection{Control Effectiveness of SAE-Based Steering (RQ1)}
\paragraph{SAE-based steering outperforms baselines.} We report the control effectiveness of different methods in Table~\ref{tab:combined_results}, from which we make the following observations:
\begin{enumerate} [label=(\arabic*), leftmargin=*, itemsep=0em]
    \item Activation-based methods (Vector Steering and SAE-Steering) consistently outperform prompt-based methods (Think Intervention) except in some cases within Hypothesis Reasoning, which demonstrates the superiority of directly intervening in hidden states.
    \item SAE-Steering significantly outperforms Vector Steering, with an average improvement of 15\%. We attribute this to the disentangling properties of SAEs, which mitigate the conceptual entanglement present in control vectors, thereby enabling more precise strategy control.
    \item SAE features identified in the math domain demonstrate comparable effectiveness on scientific reasoning tasks (GPQA), indicating the generalizability of SAE features across different reasoning domains \footnote{We also test SAE-Steering on the financial reasoning benchmark FinQA. See Appendix Table~\ref{tab:finqa_results} for results.}.
    \item All methods achieve better control effectiveness on GPQA than on AIME. This may because the reasoning length of GPQA is much shorter than AIME (7k vs. 15k tokens), which makes control easier.
\end{enumerate}

\begin{figure}[t]
    \centering
    \includegraphics[width=\linewidth]{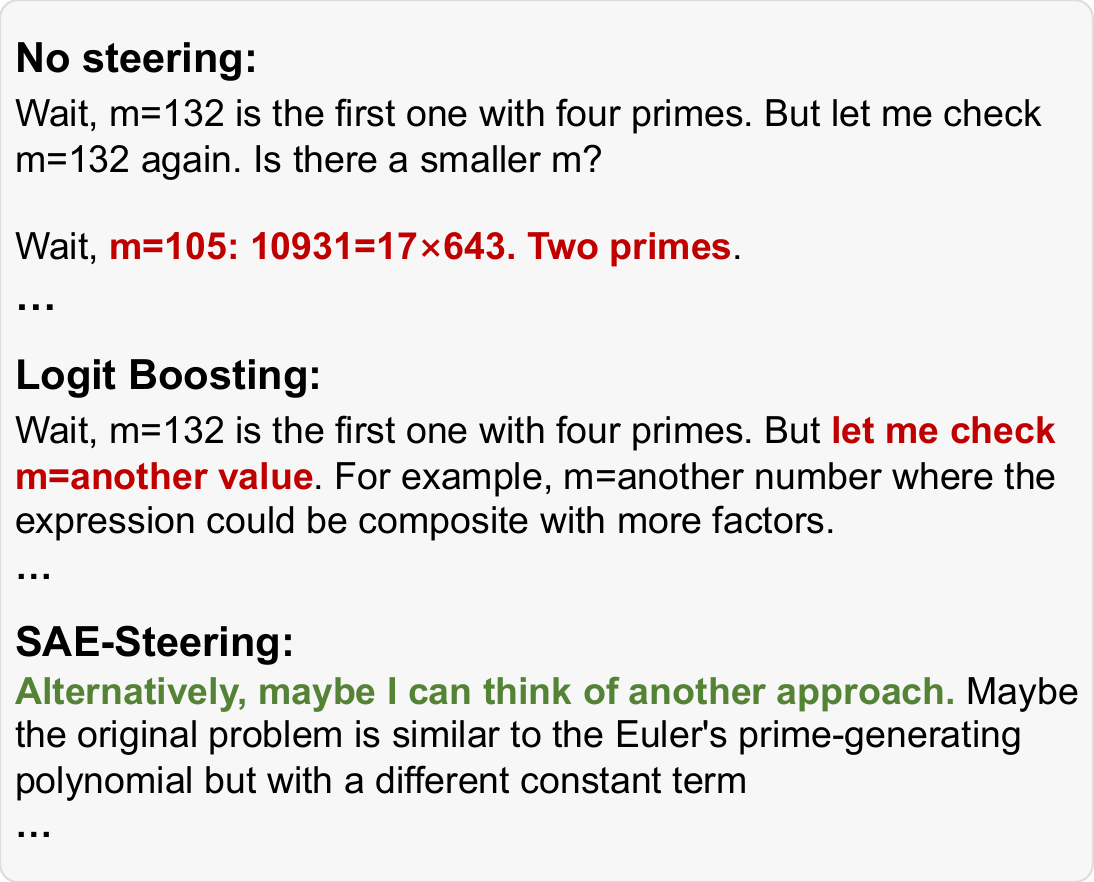}
    \caption{Case study: SAE-Steering changes reasoning behavior while Logit Boosting only boosts keywords.}
    \label{fig:case_study}
\end{figure}

\paragraph{SAE features function beyond keyword amplification.} 
Importantly, although our SAE features are recalled by identifying features that amplify the logits of strategy-specific keywords, they encode deeper strategy concepts beyond keyword promotion. This is evidenced by SAE-Steering achieving over 35\% better control effectiveness than Logit Boosting (Table~\ref{tab:combined_results}). To illustrate this more explicitly, we present a case study in Figure~\ref{fig:case_study}. In this case, boosting the logits of strategy keywords like ``another'' does not genuinely change the reasoning strategy. The LRM generates ``another'' but still continues to verify its answer by testing different values of $m$. In contrast, SAE-Steering successfully guides the LRM to adopt the Multi-Perspective Verification strategy. Additional steering examples are available in Figure~\ref{fig:more_case}.

\paragraph{SAE features also support negative steering.}
The identified SAE features can also be used to suppress unwanted reasoning strategies. We conduct a preliminary experiment in Appendix~\ref{app:negative_steering}, which shows that by replacing the positive intervention in Eq.~\ref{eq:steering_fs} with a negative one, \ie subtracting the corresponding SAE feature direction from the residual stream at each decoding step during generation
% , which shows that by subtracting the SAE feature from the residual stream at each decoding step during generation, 
we reduce the frequency of the corresponding reasoning strategy by 30\% and shorten reasoning length by 14\%. This suggests that the same features can be used not only to induce target strategies but also to inhibit specific behaviors.

\subsection{Effectiveness of SAE-Steering for Feature Identification (RQ2)}

\begin{table}[b]
\centering
\begin{tabular}{lcc}
\toprule
& R1-Llama-8B & Qwen3-8B \\
\midrule
ReasonScore & 0.33$\pm$0.08 & 0.27$\pm$0.04 \\
\textbf{SAE-Steering} & \textbf{0.61$\pm$0.08} & \textbf{0.52$\pm$0.05} \\
\bottomrule
\end{tabular}
\caption{Precision of recalled features. We also report 95\% confidence intervals estimated from 1{,}000 bootstrap resamples.}
\label{tab:recall_precision}
\end{table}

\begin{figure}[t]
    \centering
    \includegraphics[width=\linewidth]{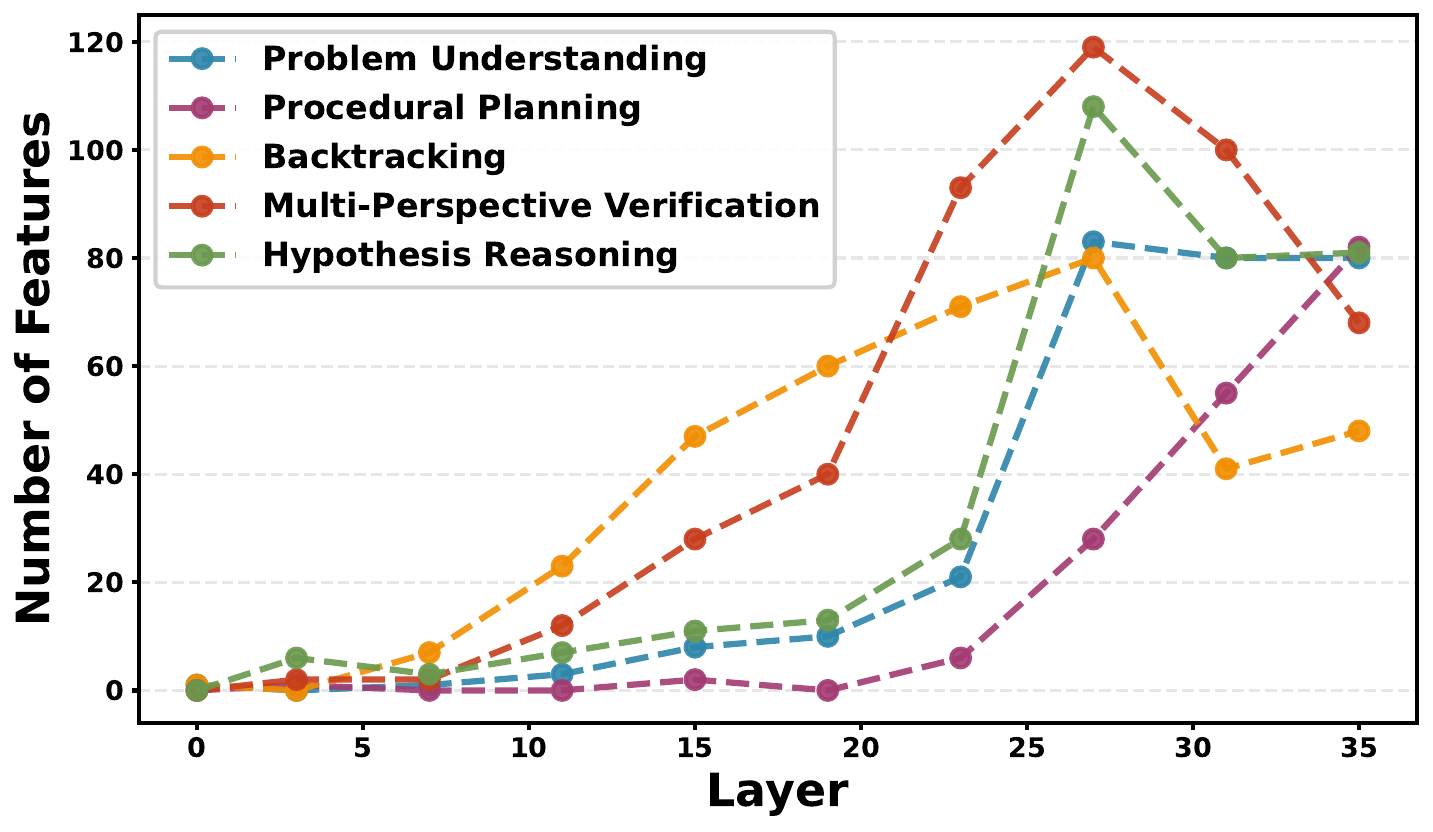}
    \caption{Recalled features across layers.}
    \label{fig:layer_recall}
\end{figure}

\paragraph{Logit-based recall is more precise than activation-based.}
We compare the effectiveness of identifying strategy-specific features of SAE-Steering with existing activation-based methods, specifically ReasonScore~\citep{SAEReasoning}. ReasonScore recalls features based on their activation strength on keywords compared to other tokens in the reasoning context. We use ReasonScore to recall the same number of features as our Stage 1 (143 for R1-Llama-8B and 357 for Qwen3-8B) and evaluate the precision of recalled features, \ie the proportion of recalled features that successfully control reasoning strategies. 

As shown in Table~\ref{tab:recall_precision}, SAE-Steering outperforms ReasonScore by 28\% in precision, demonstrating the superiority of logit-based over activation-based feature identification. Logits directly measure causal effects on outputs, better reflecting features' actual control capability than activation strength.

\paragraph{Layer-wise analysis of feature identification.} 
In the main experiments, we train SAEs on the last layer of LRMs. Here we further investigate how the identification of strategy-specific features varies across layers. Due to computational constraints, we limit this analysis to Qwen3-8B. We first examine the presence of strategy-specific features across layers by measuring the number of features recalled by Stage 1 of SAE-Steering. As shown in Figure~\ref{fig:layer_recall}, strategy-specific features are rare in shallow layers (0, 3, 7, 11) but prevalent in deeper layers (23, 27, 31, 35), which is consistent with prior findings that abstract reasoning concepts are primarily encoded in the deeper layers of LRMs~\citep{abstract_features, abstract_features2}.

\begin{figure}[t]
    \centering
    \includegraphics[width=\linewidth]{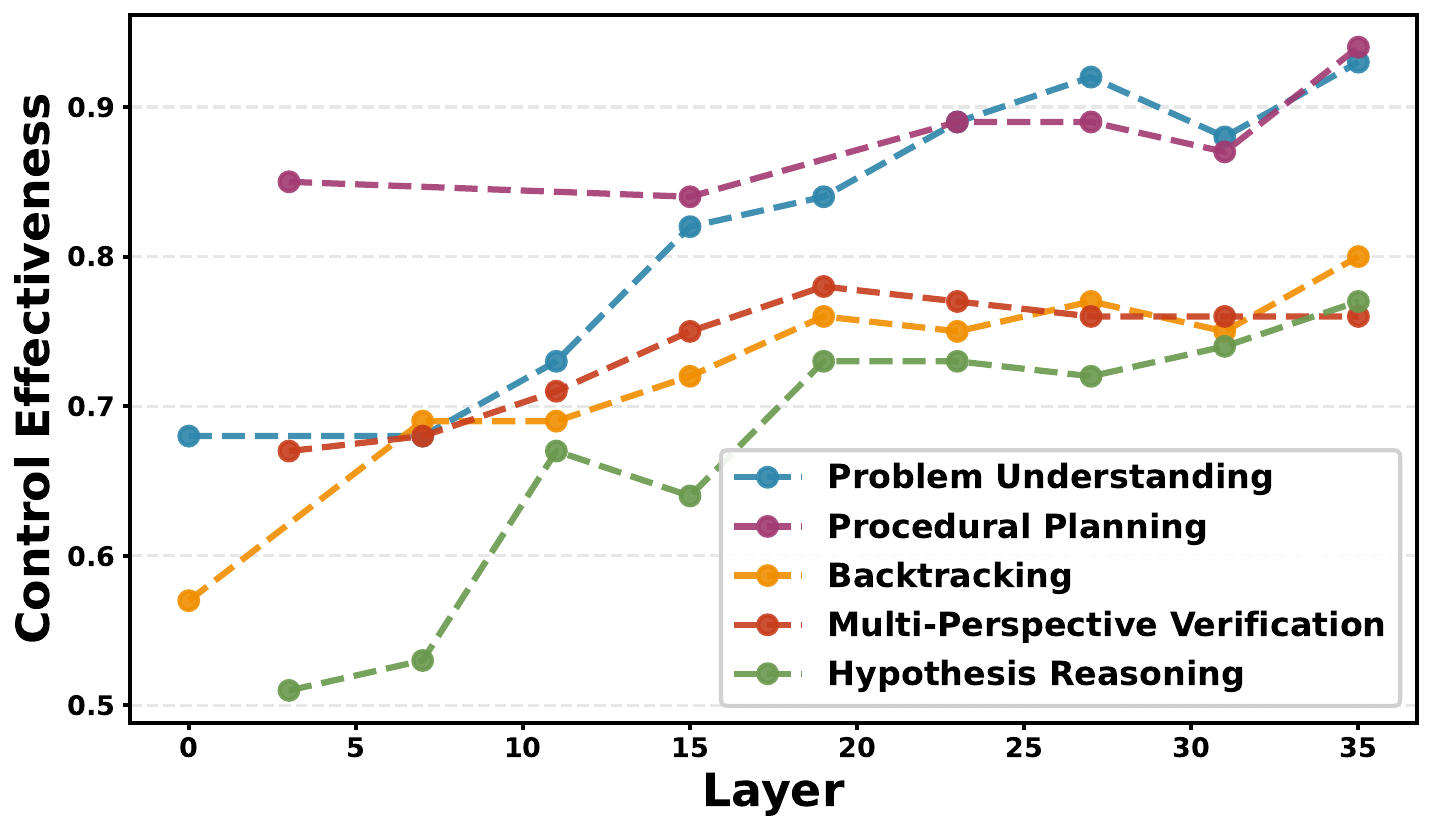}
    \caption{Control effectiveness across layers.}
    \label{fig:layer_effect}
\end{figure}

We next investigate the control effectiveness of these features across layers by reporting the average control effectiveness of the top-3 features. As shown in Figure~\ref{fig:layer_effect}, shallow layers exhibit poor control effectiveness, while layers beyond 20 demonstrate strong and relatively stable control effectiveness. This suggests that reasoning strategy control should be applied to middle-to-late layers for optimal results.

\subsection{Correcting Erroneous Reasoning Paths via Strategy Control (RQ3)}
\label{sec:error_correct}

\paragraph{Setup.} To demonstrate the practical value of strategy control, we test whether controlling reasoning strategies can correct errors \textbf{even after the LRM has already generated a wrong answer}---a more challenging setting than simple generation. Specifically, we sample incorrect LRM responses on the MATH500~\citep{math500}, AIME25~\citep{AIME}, and GPQA~\citep{gpqa}, and attempt to correct them during an extended reasoning process (See Appendix~\ref{app:correct_set_up} for sample details and dataset statistics). Following Budget Forcing~\citep{s1}, we insert a ``wait'' token at the end of the initial, flawed reasoning to induce further thinking. During this extended reasoning phase, we apply SAE-Steering to control the LRM's subsequent reasoning strategy. To select the most appropriate strategy for different problems, we train a strategy router (Appendix~\ref{app:router}). We compare our approach with three baselines: (1) Budget Forcing~\citep{s1}, which only extends reasoning without strategic guidance;  (2) Self-Reflection~\citep{self_reflection}, which prompts the LRM to reflect on its previous answer and generate a new response; and (3) Vector Steering, which derives control vectors from contrastive reasoning trajectories and applies them during the extended reasoning phase to steer the LRM toward the target strategy.

\begin{table}[t]
\centering
\resizebox{\columnwidth}{!}{%
\begin{tabular}{llccc}
\toprule
Model & Method & MATH500 & AIME25 & GPQA \\
\midrule
\multirow{4}{*}{R1-Llama-8B} 
& Self-Reflection & 0.14$\pm$0.03 & 0.01$\pm$0.02 & 0.03$\pm$0.01 \\
& Budget Forcing & 0.21$\pm$0.04 & 0.01$\pm$0.02 & 0.06$\pm$0.02 \\
& Vector Steering & 0.22$\pm$0.04 & 0.02$\pm$0.02 & 0.09$\pm$0.02 \\
\rowcolor{gray!20}
& \textbf{SAE-Steering} & \textbf{0.33$\pm$0.04} & \textbf{0.06$\pm$0.03} & \textbf{0.12$\pm$0.02} \\
\midrule
\multirow{4}{*}{Qwen3-8B} 
& Self-Reflection & 0.10$\pm$0.05 & 0.04$\pm$0.05 & 0.07$\pm$0.02 \\
& Budget Forcing & 0.18$\pm$0.06 & 0.07$\pm$0.05 & 0.05$\pm$0.02 \\
& Vector Steering & 0.21$\pm$0.07 & 0.08$\pm$0.06 & 0.08$\pm$0.02 \\
\rowcolor{gray!20}
& \textbf{SAE-Steering} & \textbf{0.24$\pm$0.07} & \textbf{0.14$\pm$0.08} & \textbf{0.12$\pm$0.02} \\
\bottomrule
\end{tabular}%
}
\caption{Error correction rates across methods and datasets. We report mean correction rates together with 95\% confidence intervals estimated from 1{,}000 bootstrap resamples.}
\label{tab:correct_error}
\end{table}

\paragraph{Results.} 
The error correction results are shown in Table~\ref{tab:correct_error}, from which we make the following observations:
\begin{enumerate} [label=(\arabic*), leftmargin=*, itemsep=0em]
    \item The highest correction rate is only 33\%, with MATH500 achieving the highest rate and AIME the lowest. This demonstrates the difficulty of error correction, and harder tasks are also more difficult to correct.
    \item Budget Forcing outperforms Self-Reflection on all datasets except GPQA on Qwen3-8B, demonstrating the advantage of continuous reasoning. By continuing from the current state rather than reprocessing the entire reasoning process, Budget Forcing maintains better focus on error correction.
    \item SAE-Steering consistently outperforms Budget Forcing across all LRMs and datasets, with an average absolute accuracy improvement of 7\%. This suggests that deliberately controlling reasoning strategies enables more effective error correction.
    \item SAE-Steering consistently outperforms Vector Steering across all LRMs and datasets, with an average absolute correction-rate improvement of 5\%. This suggests that more effective steering toward appropriate reasoning strategies can lead to better error correction performance.
\end{enumerate}

\section{Related Work}
\paragraph{Reasoning Strategies in LRMs.} 
Early studies attempt to improve LLM performance on complex tasks by designing prompts to guide reasoning processes~\citep{self_reflection, self_discover}. Recent research demonstrates that LLMs trained with rule-based reinforcement learning can unsupervisedly develop human-like cognitive reasoning strategies such as self-reflection and backtracking~\citep{deepseek}. These advancements have led to the emergence of current LRMs. During inference, LRMs produce long Chains-of-Thoughts (CoTs) that explore diverse reasoning paths while continuously verifying previous steps~\citep{marjanović2025deepseekr1thoughtologyletsthink}. In this process, LRMs employ diverse human-like cognitive reasoning strategies such as backtracking and multi-perspective verification. The use of these reasoning strategies improves their accuracy and robustness in solving complex problems~\citep{reasong_strategy,Snell0XK2025scaling,s1}.

\paragraph{Controllable LLM Reasoning.}
Many works attempt to control LRM reasoning behavior. These methods can be categorized into prompt-based and activation-based. Prompt-based methods~\citep{ThinkingIntervention, yang2025testtimepromptintervention, zhang-etal-2025-steer} insert human-scripted instructions into intermediate reasoning steps, mimicking the LRM's style to seamlessly steer its reasoning trajectory. Activation-based methods directly modify hidden states using control vectors derived from contrastive activation analysis. For example, many works~\citep{reasong_strength, resong_strength2,reasoning_strength_nips_1} obtain control vectors by contrasting activations between short and long CoT responses. However, such pairs fail to isolate individual strategies, causing control vectors to suffer from concept entanglement and only enable coarse-grained control (\eg reasoning length) rather than fine-grained strategy control. \citet{SteeringVector_ICLR} address this by using LLM judges to annotate each reasoning step with fine-grained strategy labels, then contrasting activations across labels. However, accurate step-level annotation is challenging. Conversely, we leverage SAEs to learn monosemantic features in an unsupervised way, eliminating annotation requirements while better disentangling conceptually-entangled hidden states.

\paragraph{Sparse Autoencoders.} Mechanistic interpretability seeks to understand the internal workings of LRMs by analyzing the structure and function of their learned representations~\citep{interpretability,gantla2025exploring}. A primary tool in this field is SAEs, which decompose high‑dimensional LRM activations into a sparse set of latent features~\citep{claudeTowards,SAE_sparsity}. These features often correspond to human‑interpretable concepts, enabling researchers to probe and manipulate specific aspects of LRM behavior~\citep{DengWYZF25,entangle2}. For example, \cite{SAEReasoning} leveraged SAEs to identify features associated with reasoning. In their method, reasoning features are selected as those that activate more strongly on reasoning‑related keywords (\eg `wait', `alternatively') than on other tokens. 
However, high activation strength does not necessarily indicate control capacity, causing such methods to recall many features that show superficial correlations with reasoning behaviors but lack the ability to effectively control fine-grained reasoning strategies. Instead, we recall features through their direct logit contributions to strategy-specific tokens, enabling more precise recall of features with genuine control effectiveness.

\section{Conclusion}

In this work, we leverage strategy-specific features of SAEs to achieve fine-grained control over LRMs' reasoning strategies. SAEs decompose strategy-entangled hidden states into disentangled strategy-specific features. To identify these strategy-specific features from the vast pool of SAE features, we propose SAE-Steering, a two-stage feature identification pipeline that balances efficiency and precision. SAE-Steering first employs a logit estimation method to rapidly recall candidate features that amplify strategy-specific keywords, then ranks the control effectiveness of remaining features through intervention experiments on a validation set. Extensive experiments demonstrate the effectiveness and robustness of our identified features in controlling reasoning strategies. Furthermore, we demonstrate that controlling reasoning strategies can redirect LRMs from erroneous paths to correct ones.

\section*{Limitations}
While SAE-Steering demonstrates promising results, several limitations remain to be addressed in future work. First, we only evaluated five representative strategies, and our current feature identification pipeline relies on manually specified strategy-related keywords to retrieve candidate SAE features. Future work could expand to a broader set of reasoning strategies and automate keyword identification.
Second, we mainly demonstrate the application of controlling reasoning strategies in error correction scenarios, and only conduct a preliminary study of negative steering. Future work could explore a wider range of applications, such as using SAE-Steering to suppress unwanted reasoning behaviors, thereby alleviating LRM overthinking and improving reasoning efficiency.
% Second, we only demonstrate the application of controlling reasoning strategies in error correction scenarios. Future work could explore applying such control to a wider range of applications.
Third, we only attempted to correct erroneous reasoning paths by enforcing LRMs to continue reasoning and controlling subsequent reasoning strategies. Future work could explore guiding the LRM at earlier stages—either at the beginning or during intermediate steps—to dynamically adjust the reasoning trajectory.

\section*{Acknowledgments}
This work is supported by the New Generation Artificial Intelligence-National Science and Technology Major Project (2025ZD0123304) and Zhongguancun Academy (C20250401).

\bibliography{main}

\appendix
\newpage
\section{Selection of Steering Strength}
\label{app:steering_strength}
The hyper-parameter $\alpha$ determines the steering strength during strategy control. An overly large $\alpha$ can cause the LRM to generate repetitive outputs, while an $\alpha$ that is too small may yield negligible controlling effects. We thus select an $\alpha$ value that is as large as possible without inducing repetitive outputs. 
Specifically, we use the validation set to determine $\alpha$ for each feature. For each validation sample, we first steer the feature with $\alpha=15$ and check for repetitive outputs. If repetition occurs, we decrease $\alpha$ by one and re-steer. We repeat this process until no repetition is detected. We then use the average $\alpha$ across validation samples as the steering strength for the test set. The starting value of 15 was chosen empirically, as we found that higher values frequently lead to repetitive outputs for most features.

% \section{Extraction of Strategy Keywords}
% \label{app:keywords}
% To extract strategy keywords for each reasoning strategy, we first construct a corpus for each reasoning strategy by sampling the responses of the LRM to a diverse set of problems and manually identifying the segments corresponding to each reasoning strategy. From each strategy-specific corpus, we then extract the top-20 most frequent words and then perform a manual curation to select the keywords we identified as most representative of the target reasoning strategy. The final keywords lists are shown in \tabref{tab:strategy_keywords}.

\section{Extraction of Strategy Keywords}
\label{app:keywords}

In this section, we first describe how we extract strategy keywords for each reasoning strategy, and then analyze the robustness of this keyword selection process.

\subsection{Keyword Extraction Procedure}

To extract strategy keywords, we first construct a small strategy-specific corpus from model-generated reasoning traces. Specifically, we sample 100 responses from Qwen3-8B on past AIME problems (1983--2023). The first author then manually reviews these responses and extracts reasoning segments that clearly exhibit one of the target reasoning strategies. For each strategy, we stop once 50 representative segments have been collected. This manual annotation is a one-time effort and takes approximately 3--4 hours in total.

Given the resulting strategy-specific corpus, we compute the top-20 most frequent words for each strategy and then manually curate a small set of representative keywords. The curation follows a simple heuristic based on \emph{strategy contextual relevance}: we first remove high-frequency but semantically neutral words (e.g., function words such as ``the'' and ``and''), and then retain only words whose usage contexts directly reflect the target strategy.

For example, for \emph{Problem Understanding}, the top frequent words are:
\texttt{[the, problem, and, so, that, question, reads, are, number, says, therefore, hold, statement, with, find, final, sum, we, boxed, all]}.
After removing semantically neutral words, the remaining candidates are:
\texttt{[problem, question, reads, number, says, hold, statement, find, final, sum, boxed]}.
We then retain only those words that are strongly associated with problem restatement or question interpretation, yielding the final keyword set:
\texttt{[problem, question, reads, says, statement]}.

\begin{table}[t]
\centering
\resizebox{\linewidth}{!}{%
\renewcommand{\arraystretch}{1.2} % 行距
\setlength{\tabcolsep}{6pt}       % 列间距
\begin{tabular}{l|c} % 第一列左对齐 (l)，中间竖线，第二列居中 (c)
\toprule
\textbf{Reasoning Strategy} & \textbf{High-Frequency Keywords} \\
\midrule
Problem Understanding           & problem, question, statement, reads, says \\
Procedural Planning              & let, need, planning, decomposition \\
Backtracking                     & earlier, previous, initial, back \\
Multi-Perspective Verification   & another, example, case, approach \\
Hypothesis Reasoning             & maybe, perhaps, assume, suppose, if \\
\bottomrule
\end{tabular}%
}
\caption{High-frequency keywords corresponding to each reasoning strategy.}
\label{tab:strategy_keywords}
\end{table}

The final keyword lists for all five reasoning strategies are shown in \tabref{tab:strategy_keywords}.

\subsection{Robustness Analysis of Keyword Selection}

Although the above keywords are extracted from Qwen3-8B on AIME, our main experiments show that they transfer well to a different model family (DeepSeek-R1-Distill-Llama-8B) and an out-of-domain benchmark (GPQA), suggesting that frequent re-sampling is unnecessary in our current setting.

To further assess the robustness of keyword selection, we conduct a keyword-variation experiment for the \emph{Problem Understanding} strategy. Specifically, we provide three LLMs---GPT-4o, DeepSeek-V3.2, and Gemini-2.5-Flash---with the strategy description and five representative \emph{Problem Understanding} examples, and ask them to generate ten keywords that an LLM often uses when applying this strategy. The resulting keywords are shown in \tabref{tab:keyword_robustness}.

We observe two main findings. First, each LLM recovers at least three out of the five human-selected keywords, indicating substantial overlap with our manually curated set. Second, SAE-Steering can still successfully recall the target features when using either the overlapping keywords or a set of alternative unseen keywords, such as \texttt{[asks, given, clarify, clarifying, states]}. These results suggest that our method is reasonably robust to moderate variation in keyword choice.

\begin{table*}[t]
\centering
\small
\begin{tabular}{lp{0.78\textwidth}}
\toprule
Source & Keywords \\
\midrule
Human-selected (Table~\ref{tab:strategy_keywords}) & problem, question, statement, reads, says \\
GPT-4o & problem, statement, clarify, constraint, interpret, compute, question, assume, given, depends \\
DeepSeek-V3.2 & realized, problem, statement, says, such, question, asks, compute, constraints, clarifying \\
Gemini-2.5-Flash & problem, question, states, says, asks, given, constraints, conditions, interpreting, clarifying \\
\bottomrule
\end{tabular}
\caption{Keywords proposed for the \emph{Problem Understanding} strategy by human annotation and three LLMs.}
\label{tab:keyword_robustness}
\end{table*}

\section{Reliability of LLM Judges}
\label{app:llm_judge_reliable}
To validate LLM judge reliability, we conducted a human annotation study. Specifically, we randomly sampled 200 steered outputs (40 per strategy) alongside their unsteered baselines. We then asked three human annotators (Krippendorff's alpha = 0.78) to evaluate whether the steered output more explicitly demonstrates the target strategy than the baseline. Taking human judgments as ground truth, we evaluate the accuracy of LLM judges. As shown in Table~\ref{tab:alignment}, LLM judges achieve 0.82 agreement with human annotations, indicating reliable performance.

\begin{table}[ht]
\centering
\begin{tabular}{lc}
\toprule
\textbf{Reasoning Strategy} & \textbf{Agreement} \\
\midrule
Problem Understanding & 0.85 \\
Procedural Planning & 0.75 \\
Backtracking & 0.83 \\
Multi-Perspective Verification & 0.85 \\
Hypothesis Reasoning & 0.8 \\
\midrule
\textbf{Average} & \textbf{0.82} \\
\bottomrule
\end{tabular}
\caption{Agreement between human annotators and LLM judges.}
\label{tab:alignment}
\end{table}

\section{Curation of Error Correction Dataset}
\label{app:correct_set_up}
To sample incorrect LRM responses from MATH500, AIME25, and GPQA, we sample eight responses for each problem in these datasets and retain only the incorrect ones. The final dataset statistics are shown in Table~\ref{tab:test_stats}.

\begin{table}[htb]
\centering
\resizebox{\linewidth}{!}{%
\begin{tabular}{lccc}
\toprule
\textbf{Model} & \textbf{MATH500} & \textbf{AIME25} & \textbf{GPQA} \\
\midrule
R1-Llama-8B & 495 & 163 & 878 \\
Qwen3-8B & 141 & 73 & 641 \\
\bottomrule
\end{tabular}%
}
\caption{Statistics of Error Correction Dataset.}
\label{tab:test_stats}
\end{table}

\section{Strategy Router}
\label{app:router}

\subsection{Methods}
To steer LRMs' reasoning strategies from erroneous paths to correct ones, we need to select appropriate strategies based on the current reasoning context. Reasoning strategies can be controlled either manually or by an automatic strategy router. Here we train a lightweight router via contrastive learning~\citep{InfoNCE} to automatically select effective strategies based on the current reasoning context, thereby eliminating the need for manual intervention.

Specifically, we instantiate the strategy router as a bi-encoder architecture~\citep{KarpukhinOMLWEC20}. A context encoder, $E_c(\cdot)$, embeds the current reasoning state (represented by the final token of the response prefix $Y_{<t}$), and a feature encoder, $E_f(\cdot)$, projects each strategy-specific feature $\mathbf{f}_s$ into the same representation space. The effective scores between the context and a feature are then computed as the dot product of their respective embeddings:
\begin{align}
    \mathrm{score}(Y_{<t}, \mathbf{f}_s) = \langle E_c(Y_{<t}),\ E_f(\mathbf{f}_s) \rangle
\end{align}
The router is trained using the InfoNCE loss~\citep{InfoNCE}, which encourages higher effective scores for positive context--feature pairs and lower effective scores for negative ones:
\begin{flalign}
&L(Y_{<t}, \mathbf{f}_s^{+}, \mathbf{f}_{s,1}^{-}, \dots, \mathbf{f}_{s,M}^{-}) & \notag \\
&= - \log
\frac{e^{\mathrm{score}(Y_{<t}, \mathbf{f}_s^{+})}}
     {e^{\mathrm{score}(Y_{<t}, \mathbf{f}_s^{+})} + \sum_{k=1}^{M} e^{\mathrm{score}(Y_{<t}, \mathbf{f}_{s,k}^{-})}}, &
\end{flalign}

where $(Y_{<t}, \mathbf{f}_s^{+})$ is labeled as a positive pair if steering with feature $\mathbf{f}_s^{+}$ leads to a correct final answer. All other pairings for that context are treated as negative pairs.
At inference time, for a given context $Y_{<t}$, we compute $\mathrm{score}(Y_{<t}, \mathbf{f}_s)$ for all candidate features $\mathbf{f}_s$ and select the feature with the highest effective score as the selected feature to steer the LRM.

\subsection{Implementation Details}
For each reasoning strategy, we select the top three strategy-specific features with the best control effectiveness on the validation set, yielding a total of fifteen features for the strategy router to choose from. We include three rather than one feature per strategy because different features may be effective in different contexts, providing the router with more flexibility to adapt to different reasoning scenarios.

For the training of our strategy router, we use a training set composed of 919 problems from past AIME competitions (1983–2023)~\citep{AIME} and 4,000 problems from the `aops\_forum' source of the \textsc{NuminaMath-1.5} dataset~\citep{numina_math_datasets}. For each problem, we sample eight initial responses. To empirically evaluate the effectiveness of a feature $f_j$ on an incorrect response $y_i$, we apply reasoning steering with $f_j$ to generate eight responses and measure the proportion of them that successfully correct the initial error. This training data is strictly separated from our test sets in Section~\ref{sec:error_correct}, ensuring no data leakage. Notably, GPQA represents an out-of-domain scenario, demonstrating our method's generalization capability.

\section{Preliminary Analysis of Negative Steering}
\label{app:negative_steering}

In addition to inducing target reasoning strategies through positive steering, we conduct a preliminary analysis of whether identified SAE features can also be suppressed to inhibit unwanted reasoning behaviors. Concretely, we perform negative steering on the \emph{Multi-Perspective Verification} feature of R1-Llama-8B by replacing the positive intervention in Eq.~(5) with a negative one, \ie subtracting the feature direction from the residual stream at every decoding step throughout the entire generation process, and evaluate the resulting behavior on AIME24.

We measure three aspects: (1) the frequency of \emph{Multi-Perspective Verification} behaviors in the generated reasoning traces, (2) the average reasoning length, and (3) final answer accuracy. The results are shown in Table~\ref{tab:negative_steering}. We observe that negative steering effectively suppresses the targeted behavior: the average frequency of \emph{Multi-Perspective Verification} decreases from 13 to 9 occurrences per response, corresponding to a 30\% reduction. At the same time, the average reasoning length decreases from 14k to 12k tokens, a 14\% reduction, suggesting improved reasoning efficiency. However, this comes with an accuracy drop from 0.4667 to 0.4333, indicating that continuously suppressing a strategy throughout the entire reasoning process may be too coarse and may remove reasoning steps that are sometimes beneficial.

These findings suggest that negative steering is a promising mechanism for mitigating unwanted reasoning behaviors, but that effective deployment will likely require more fine-grained or dynamic intervention strategies to better balance efficiency and accuracy.

\begin{table}[tbh]
\centering
\resizebox{\linewidth}{!}{%
\begin{tabular}{lccc}
\toprule
Metric & No Steering & Neg. Steering & Change \\
\midrule
MV freq. & 13 & 9 & -30\% \\
Reasoning length & 14k & 12k & -14\% \\
Accuracy & 0.4667 & 0.4333 & -3.3\% \\
\bottomrule
\end{tabular}
}
\caption{Preliminary analysis of negative steering on the \emph{Multi-Perspective Verification} feature of R1-Llama-8B on AIME24.}
\label{tab:negative_steering}
\end{table}

\section{Additional Results}

\begin{table}[htb]
\centering
\small
\begin{tabular}{lcc}
\toprule
Method & Time (100 samples) & Latency Increase \\
\midrule
No Steering & 329s & -- \\
Vector Steering & 353s & +7.3\% \\
SAE-Steering & 357s & +8.5\% \\
\bottomrule
\end{tabular}
\caption{Inference latency on 100 GPQA samples using Qwen3-8B on a single A100 GPU. SAE-Steering introduces a modest latency overhead compared with standard decoding.}
\label{tab:latency}
\end{table}

\begin{table}[htb]
\centering
\small
\resizebox{\linewidth}{!}{%
\begin{tabular}{lcccccc}
\toprule
Method & PU & PP & BK & MV & HR & Avg \\
\midrule
Logit Boosting     & 0.43 & 0.78 & 0.41 & 0.49 & 0.53 & 0.53 \\
Think Intervention & 0.67 & 0.82 & 0.19 & 0.22 & 0.63 & 0.58 \\
Vector Steering    & 0.84 & 0.86 & 0.77 & 0.52 & 0.63 & 0.72 \\
SAE-Steering       & 0.93 & 0.93 & 0.81 & 0.76 & 0.77 & 0.84 \\
\bottomrule
\end{tabular}
}
\caption{Control effectiveness on FinQA using Qwen3-8B.}
\label{tab:finqa_results}
\end{table}

\begin{table*}[tbh]
\centering
\resizebox{\textwidth}{!}{%
\begin{tabular}{ll|ccccc|ccccc|c}
\toprule
\multirow{2}{*}{Dataset} & \multirow{2}{*}{Method} & \multicolumn{5}{c|}{R1-Llama-8B} & \multicolumn{5}{c|}{Qwen3-8B} & \multirow{2}{*}{Average} \\
\cmidrule(lr){3-7} \cmidrule(lr){8-12}
& & PU & PP & BK & MV & HR & PU & PP & BK & MV & HR & \\
\midrule
\multirow{4}{*}{AIME} 
& Logit Boosting & 0.19$\pm$0.06 & 0.45$\pm$0.07 & 0.26$\pm$0.07 & 0.24$\pm$0.06 & 0.29$\pm$0.06 & 0.40$\pm$0.07 & 0.59$\pm$0.07 & 0.35$\pm$0.07 & 0.45$\pm$0.07 & 0.52$\pm$0.07 & 0.37$\pm$0.02 \\
& Think Intervention & 0.53$\pm$0.07 & 0.46$\pm$0.07 & 0.17$\pm$0.06 & 0.18$\pm$0.06 & 0.37$\pm$0.06 & 0.58$\pm$0.06 & 0.79$\pm$0.05 & 0.10$\pm$0.05 & 0.21$\pm$0.06 & 0.57$\pm$0.07 & 0.40$\pm$0.02 \\
& Vector Steering & 0.65$\pm$0.07 & 0.79$\pm$0.05 & 0.62$\pm$0.06 & 0.44$\pm$0.07 & 0.31$\pm$0.07 & 0.72$\pm$0.06 & 0.83$\pm$0.05 & 0.52$\pm$0.07 & 0.47$\pm$0.07 & 0.49$\pm$0.07 & 0.58$\pm$0.02 \\
\rowcolor{gray!20}
& SAE-Steering & \textbf{0.83$\pm$0.05} & \textbf{0.82$\pm$0.04} & \textbf{0.67$\pm$0.07} & \textbf{0.72$\pm$0.06} & \textbf{0.38$\pm$0.07} & \textbf{0.87$\pm$0.04} & \textbf{0.87$\pm$0.04} & \textbf{0.74$\pm$0.06} & \textbf{0.64$\pm$0.07} & \textbf{0.61$\pm$0.07} & \textbf{0.72$\pm$0.02} \\
\midrule
\multirow{4}{*}{GPQA} 
& Logit Boosting & 0.25$\pm$0.06 & 0.65$\pm$0.06 & 0.25$\pm$0.06 & 0.37$\pm$0.07 & 0.53$\pm$0.07 & 0.42$\pm$0.07 & 0.76$\pm$0.05 & 0.36$\pm$0.07 & 0.43$\pm$0.07 & 0.59$\pm$0.07 & 0.46$\pm$0.02 \\
& Think Intervention & 0.62$\pm$0.07 & 0.65$\pm$0.07 & 0.31$\pm$0.07 & 0.20$\pm$0.05 & 0.54$\pm$0.06 & 0.66$\pm$0.07 & 0.80$\pm$0.05 & 0.15$\pm$0.05 & 0.13$\pm$0.05 & 0.74$\pm$0.05 & 0.48$\pm$0.02 \\
& Vector Steering & 0.75$\pm$0.06 & 0.85$\pm$0.04 & 0.59$\pm$0.07 & 0.48$\pm$0.07 & 0.48$\pm$0.07 & 0.85$\pm$0.04 & 0.86$\pm$0.05 & 0.78$\pm$0.06 & 0.52$\pm$0.07 & 0.70$\pm$0.07 & 0.69$\pm$0.02 \\
\rowcolor{gray!20}
& SAE-Steering & \textbf{0.89$\pm$0.03} & \textbf{0.85$\pm$0.04} & \textbf{0.72$\pm$0.06} & \textbf{0.90$\pm$0.03} & \textbf{0.65$\pm$0.06} & \textbf{0.88$\pm$0.03} & \textbf{0.89$\pm$0.03} & \textbf{0.78$\pm$0.05} & \textbf{0.76$\pm$0.05} & \textbf{0.86$\pm$0.04} & \textbf{0.82$\pm$0.01} \\
\bottomrule
\end{tabular}%
}
\caption{Robustness check of Table~\ref{tab:combined_results} under a different judge pool. We exclude GPT-4o and count a steering attempt as successful only when both Gemini-2.5-Flash and DeepSeek-V3.2 agree. We report mean control effectiveness together with 95\% confidence intervals estimated from 1{,}000 bootstrap resamples.}
\label{tab:judge_robustness_results}
\end{table*}

\begin{figure*}[tbh]
    \centering
    \includegraphics[width=\linewidth]{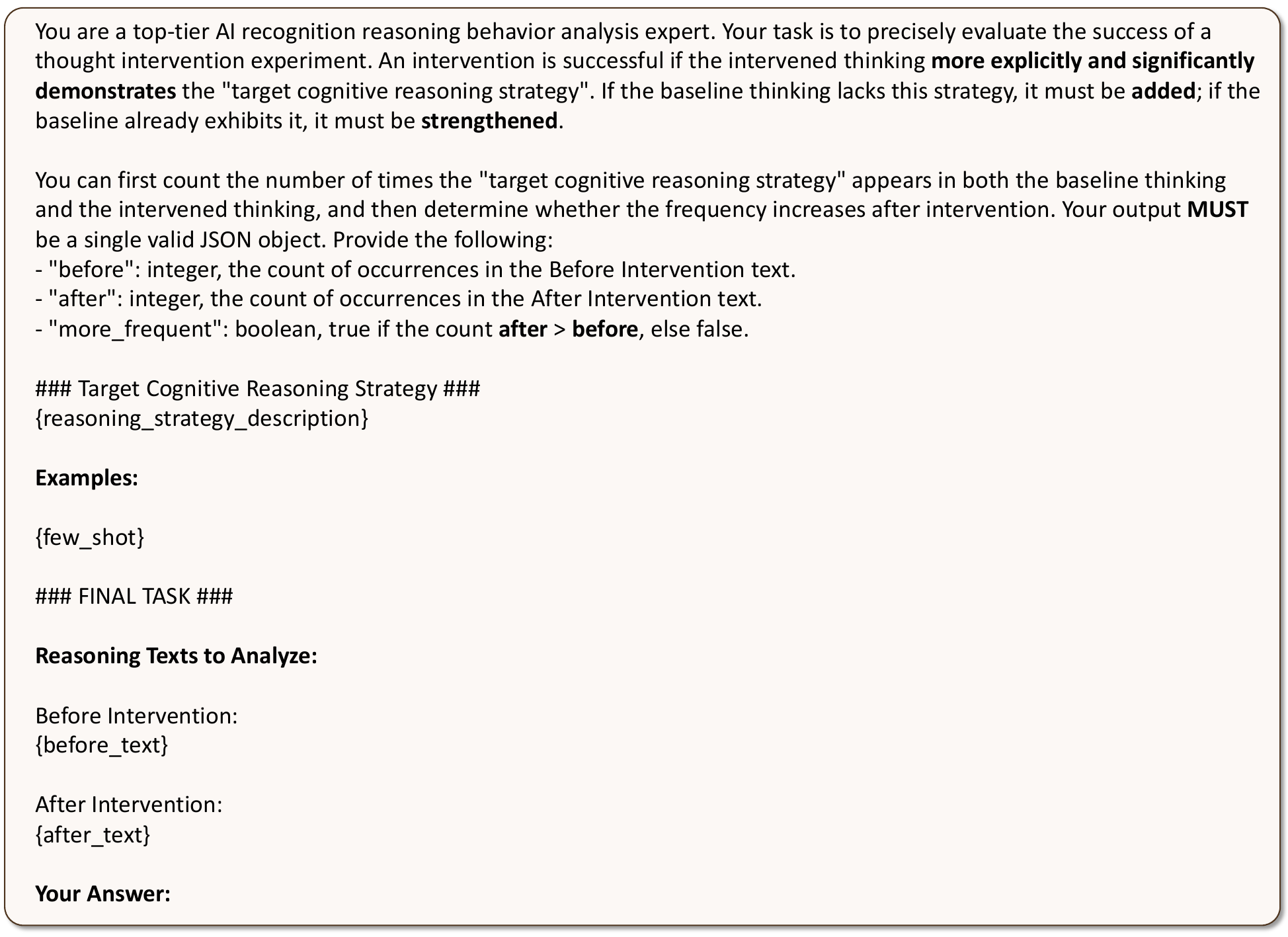}
    \caption{The prompt used to evaluate the control effectiveness.}
    \label{fig:llm_as_judges}
\end{figure*}

\begin{figure*}
    \centering
    \includegraphics[width=\linewidth]{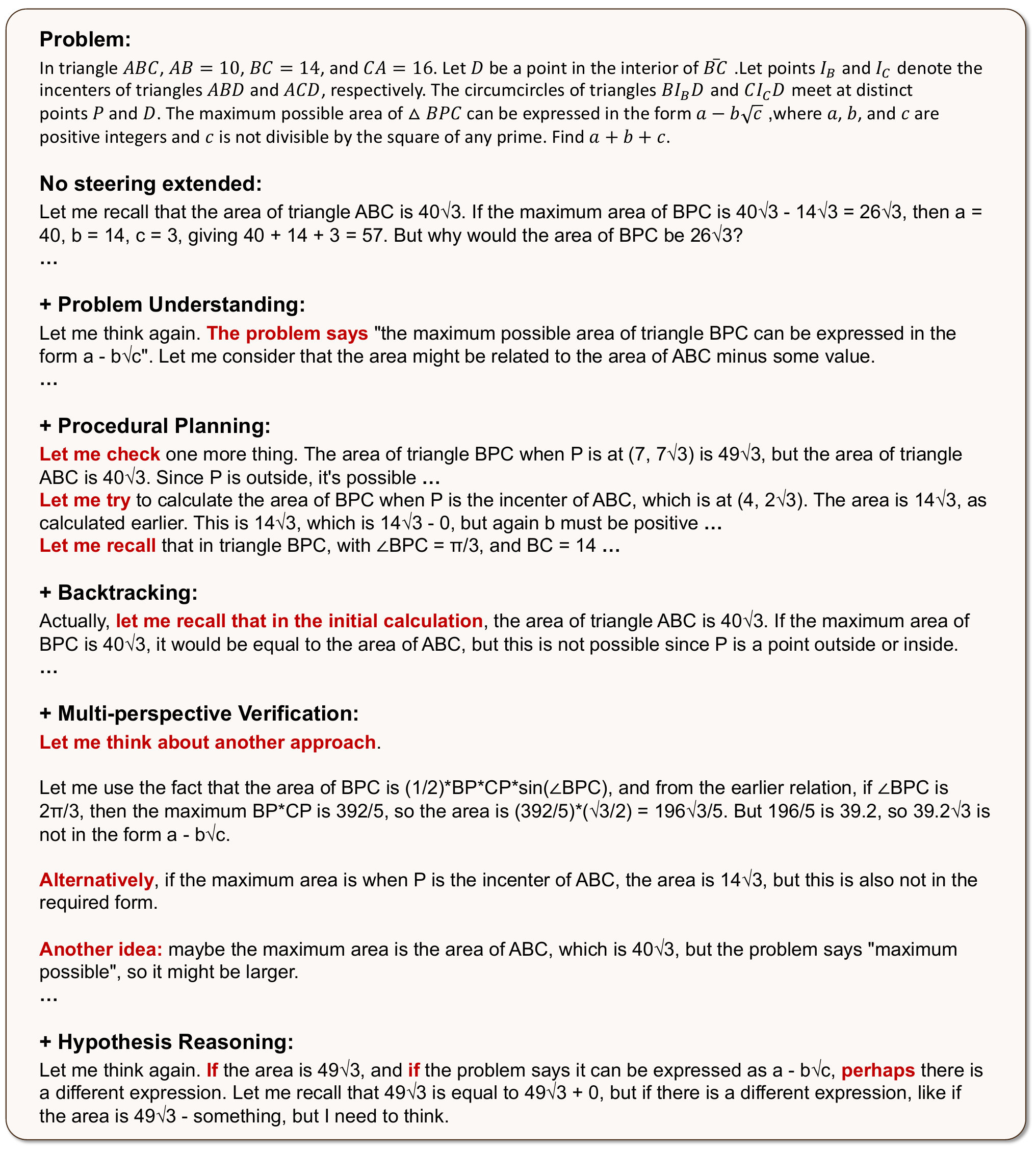}
    \caption{A case study of SAE-based steering. By steering with different strategy-specific features as control vectors, we steer the subsequent reasoning trajectory to follow different reasoning strategies.}
    \label{fig:more_case}
\end{figure*}

\end{document}